\documentclass{article}

\usepackage[utf8]{inputenc} % allow utf-8 input
\usepackage[T1]{fontenc}    % use 8-bit T1 fonts
\usepackage{hyperref}       % hyperlinks
\usepackage{url}            % simple URL typesetting
\usepackage{booktabs}       % professional-quality tables
\usepackage{amsfonts}       % blackboard math symbols
\usepackage{nicefrac}       % compact symbols for 1/2, etc.
\usepackage{microtype}      % microtypography

\usepackage{amsmath}
\usepackage[pdftex]{graphicx}
\usepackage[flushleft]{threeparttable}

\usepackage{graphicx}

\usepackage{multirow}

\DeclareMathOperator*{\argmin}{arg\,min}

\title{A Probabilistic Representation of Deep Learning}

\author{%
  Xinjie Lan, Kenneth E. Barner\\
  Department of Electrical and Computer Engineering\\
  University of Delaware\\
  Newark, DE 19713 \\
  \texttt{lxjbit@udel.edu} \\
}

\begin{document}

\maketitle

\begin{abstract}
In this work, we introduce a novel probabilistic representation of deep learning, which provides an explicit explanation for the Deep Neural Networks (DNNs) in three aspects: (i) neurons define the energy of a Gibbs distribution; (ii) the hidden layers of DNNs formulate Gibbs distributions; and (iii) the whole architecture of DNNs can be interpreted as a Bayesian neural network.
Based on the proposed probabilistic representation, we investigate two fundamental properties of deep learning: hierarchy and generalization. 
First, we explicitly formulate the hierarchy property from the Bayesian perspective, namely that some hidden layers formulate a prior distribution and the remaining layers formulate a likelihood distribution.
Second, we demonstrate that DNNs have an explicit regularization by learning a prior distribution and the learning algorithm is one reason for decreasing the generalization ability of DNNs. 
Moreover, we clarify two empirical phenomena of DNNs that cannot be explained by traditional theories of generalization. 
Simulation results validate the proposed probabilistic representation and the insights into these properties of deep learning based on a synthetic dataset.

\end{abstract}

\section{Introduction}

%Deep learning is a subset of machine learning algorithms that construct the Deep Neural Networks (DNNs) to solve complex problems \cite{CNN-Nature}.
%Though it has achieved great success in various fields,  such as speech recognition \cite{ANN-GMM} and image classification \cite{CNN}, the working mechanism of deep learning is still not convincingly explainable and DNNs have been regarded as "black boxes" \cite{DNN_blackbox}.

Based on an underlying premise that DNNs establish a complex probabilistic model \cite{posterior1, pearl, cnn-posterior, posterior4}, numerous theories, such as representation learning \cite{representation-dl, Goodfellow-et-al-2016, DRMM}, information bottleneck \cite{Soatto1, DNN-Bottleneck, IP-argue, DNN-information}, %variational inference \cite{gd_vr, posterior1, ml_vi, cnn-posterior, posterior4}, 
have been proposed  to explore the working mechanism of deep learning.
Though the proposed theories reveal some important properties of deep learning, such as hierarchy \cite{representation-dl, Goodfellow-et-al-2016} and sufficiency \cite{Soatto1, DNN-information},
a fundamental problem is that the underlying premise is still not explicitly formulated. 

In the context of probabilistic modeling for deep learning, most previous works focus on finding a probabilistic model to explain a single hidden layer of DNNs.
It is known that every hidden layer of the Deep Boltzmann Machine (DBM) is equivalent to the restricted Boltzmann distribution \cite{Boltzmann_machine, deep-BM}.
Some works demonstrate that a convolutional layer can be explained as an undirected probabilistic graphical model, namely the Markov Random Fields (MRFs) \cite{xinjie2, CRF_RNN}.
In addition, the softmax layer is proved to be a discrete Gibbs distribution \cite{Gibbs_softmax}.
However, there are still some hidden layers, such as fully connected layer, without clearly probabilistic explanation.
Although it is known that DNNs stack hidden layers in a hierarchical way \cite{representation-dl, Boltzmann_machine, DRMM, DMFA}, establishing an explicitly probabilistic explanation for the whole architecture of DNNs has never been attempted successfully.
In summary, we still don't know what is the exactly probabilistic model corresponding to DNNs.

The obscurity of the premise impedes clearly formulating some important properties of deep learning.
First, we are still unclear what is the exact principle of assembling various hidden layers into a hierarchical neural network for a specific application \cite{representation-dl, Goodfellow-et-al-2016}.
Second, though DNNs achieve great generalization performance, we cannot convincingly formulate the {generalization} property of DNNs based on traditional complexity measures, e.g., the VC dimension \cite{VC-dimension} and the uniform stability \cite{uniform_stability}.
Recent works claim that DNNs perform implicit {regularization} by the Stochastic Gradient Descent (SGD) \cite{gd_vr, generalization_regularization1}, but they cannot clarify some empirical phenomena of DNNs presented in \cite{regularization5, generalization_regularization}.

To establish an explicitly probabilistic premise for deep learning, we introduce a novel probabilistic representation of deep learning based on the Markov chain \cite{it_book, DNN-information} and the Energy Based Model (EBM) \cite{Geman, energy_learning}.
More specifically, we provide an explicitly probabilistic explanation for DNNs in three aspects: (i) neurons define the energy of a Gibbs distribution; (ii) hidden layers formulate Gibbs distributions; and (iii) the whole architecture of DNNs can be interpreted as a Bayesian Hierarchical Model (BHM).
To the best of our knowledge, this is the first probabilistic representation that can comprehensively interpret every component and the whole architecture of DNNs.

Based on the proposed probabilistic representation, we provide novel insights into two properties of DNNs: {hierarchy} and {generalization}. 
Above all, we explicitly formulate the hierarchy property of deep learning from the Bayesian perspective, namely that the hidden layers close to the training dataset $\boldsymbol{x}$ model a prior distribution $q(\boldsymbol{X})$ and the remaining layers model a likelihood distribution $q(\boldsymbol{Y|X})$ for the training labels $\boldsymbol{y}$.
Second, unlike previous work claiming that DNNs perform implicit {regularization} by SGD \cite{generalization_regularization1, generalization_regularization}, we demonstrate that DNNs have an explicit {regularization} by learning $q(\boldsymbol{X})$ based on the Bayesian regularization theory \cite{bayesian_regularization} and prove that SGD is a reason for decreasing the generalization ability of DNNs from the perspective of the variational inference \cite{gd_vr, SVI}.
%$\text{KL}[p(\boldsymbol{X})||q(\boldsymbol{X})]$ can be a theoretical complexity measure for explaining the {generalization} of deep learning, where $p(\boldsymbol{X})$ is the truly prior distribution of the training dataset.

Moreover, we clarify two empirical phenomena of DNNs that are inconsistent with traditional theories of generalization \cite{generalization_regularization1}.
First, increasing the number of hidden units can decrease the generalization error but not result in overfitting even in an over-parametrized DNN \cite{regularization5}.
That is because more hidden units enable DNNs to formulate a better prior distribution $q(\boldsymbol{X})$ to regularize the likelihood distribution $q(\boldsymbol{Y|X})$, thereby guaranteeing the generalization performance.  
Second, DNNs can achieve zero training error but high generalization error for random labels \cite{generalization_regularization}.
We demonstrate the DNN still have good generalization performance in terms of learning an accurate prior distribution $q(\boldsymbol{X})$.
The high generalization error is due to the fact that it is impossible for arbitrary DNNs to classify random labels because it can only model two dependent random variables. %modeling the connection between two independent random variables.
%SGD could decrease the \textit{generalization} ability of DNNs because it only minimizes the distance between the $q(\boldsymbol{Y|X})$ derived from DNNs and the truly conditional distribution $p(\boldsymbol{Y|X})$ but DNNs actually model a joint distribution $q(\boldsymbol{Y, X}) = q(\boldsymbol{Y|X})q(\boldsymbol{X})$. 
%It means that SGD merely learns the prior distribution $q(\boldsymbol{X})$ that helps DNNs deriving the $q(\boldsymbol{Y|X})$ close to $p(\boldsymbol{Y})$, but the learned  the one that is close to the truly prior distribution $p(\boldsymbol{X})$.
%which answers some fundamental but still unsolved problems of \textit{generalization}, e.g., 
%Third, we review the \textit{sufficiency} property of deep learning derived from IB theory \cite{DNN-Bottleneck, DNN-information}, and demonstrate that \textit{sufficiency} only holds for the output layer of DNNs but not for every hidden layer.

\section{Related work}

\subsection{Energy based model} 

The Energy Based Model (EBM) describes the dependencies within the input $\boldsymbol{x}$ by associating an energy to each configuration of $\boldsymbol{x}$ \cite{energy_learning}. 
We commonly formulate EBM as a Gibbs distribution
\begin{equation} 
\label{Gibbs} 
{\textstyle
p(\boldsymbol{x}; \boldsymbol{\theta}) = \frac {1}{Z(\boldsymbol{\theta})}\text{exp}[-E(\boldsymbol{x; \theta})]\text{,}
}
\end{equation}
where $E(\boldsymbol{x; \theta})$ is the energy function, $Z(\boldsymbol{\theta}) = \sum_{\boldsymbol{x}}{\text{exp}[-E(\boldsymbol{x; \theta})]}$ is the partition function, and $\boldsymbol{\theta}$ denote all parameters.
A classical example of EBM in deep learning is the Boltzmann machine \cite{deep-BM}.
In particular, the Gibbs distribution belongs to the exponential family \cite{exponential} and can be expressed as
\begin{equation} 
\label{exponential} 
{\textstyle
p(\boldsymbol{x}; \boldsymbol{\theta}) = \text{exp}[\langle \boldsymbol{t(x)}, \boldsymbol{\theta} \rangle - F(\boldsymbol{\theta})]\text{,}
}
\end{equation}
where $\boldsymbol{t(x)}$ is the sufficient statistics for $p(\boldsymbol{x}; \boldsymbol{\theta})$, $F(\boldsymbol{\theta}) = logZ(\boldsymbol{\theta})$ is called the log-normalizer, and $\langle \cdot, \cdot \rangle$ denotes the inner product \cite{it_book}. 
We can derive that $E(\boldsymbol{x; \theta})$ is a sufficient statistics for $p(\boldsymbol{x}; \boldsymbol{\theta})$ as well because $E(\boldsymbol{x; \theta}) = - \langle \boldsymbol{t(x)}, \boldsymbol{\theta} \rangle$.
In addition, {conjugacy} is an important property of Gibbs distribution, which indicates that the posterior distribution would be a Gibbs distribution if the prior and the likelihood distributions are both Gibbs distributions \cite{conjugate_gibbs, Conjugate}.
% and (iii) $ E_{\boldsymbol{\theta}}[\boldsymbol{t(x)}] = \frac{\partial F(\boldsymbol{\theta})}{\partial \boldsymbol{\theta}}$.

\subsection{Stochastic variational inference} 
As a dominant paradigm for posterior inference $p(H|E) \propto {p(E|H)\cdot p(H)}$, the variational inference converts the inference problem into an optimization problem \cite{VI}, where the prior distribution $p(H)$ is the probability of arbitrary hypothesis $H$ with respect to the observation $E$ and the likelihood distribution $p(E|H)$ is the probability of $E$ given $H$.
More specifically, variational inference posits a family of approximate distributions $\mathcal{Q}$ and aims to find a distribution $q^{*}(H)$ that minimizes the Kullback-Leibler (KL) divergence between $p(H|E)$ and $q(H)$.
\begin{equation} 
\label{VI} 
q^{*}(H) = \argmin_{q \in \mathcal{Q}} \text{KL}(p(H|E)||q(H))
\end{equation}

A typical method to solve the above optimization problem is the stochastic variational inference \cite{SVI}, which iteratively optimizes each random variable in $H$ based on the samples of $E$ while holding other random variables fixed until achieving a local minimum of $\text{KL}(p(H|E)||q(H))$.
In particular, previous works prove that SGD performs variational inference during training DNNs \cite{gd_vr, ml_vi}.

\section{The probabilistic representation of deep learning}

%To facilitate subsequent discussions, 
We assume that $p_{\boldsymbol{\theta}}(\boldsymbol{X}, \boldsymbol{Y}) = p(\boldsymbol{Y|X})p(\boldsymbol{X})$ is an unknown joint distribution between $\boldsymbol{X}$ and $\boldsymbol{Y}$, where $p(\boldsymbol{X})$ describes the prior knowledge of $\boldsymbol{X}$, $p(\boldsymbol{Y|X})$ describes the statistical connection between $\boldsymbol{X}$ and $\boldsymbol{Y}$, and $\boldsymbol{\theta}$ denote the parameters of $p_{\boldsymbol{\theta}}(\boldsymbol{X}, \boldsymbol{Y})$. 
In addition, a training dataset $\boldsymbol{\mathcal{D}} = \{(\boldsymbol{x}_n, \boldsymbol{y}_n)| \boldsymbol{x}_n \in \boldsymbol{R}^{S}, \boldsymbol{y}_n \in \boldsymbol{R}^{L}\}_{n=1}^{N}$ is composed of i.i.d. samples generated from $p_{\theta}(\boldsymbol{X}, \boldsymbol{Y})$. 
A neural network with $I$ hidden layers is denoted as $\text{DNN} = \{\boldsymbol{x; f_1; ...; f_I; f_Y}\}$ and trained by $\boldsymbol{\mathcal{D}}$.
$\boldsymbol{F_i}$ is the random variable for the hidden layer $\boldsymbol{f_i}$, and $\boldsymbol{f_Y}$ is the estimation of the distribution $p(\boldsymbol{Y})$.

\begin{figure}[t]
\centering
\begin{minipage}[b]{0.68\linewidth}
\centerline{\includegraphics[scale=0.13]{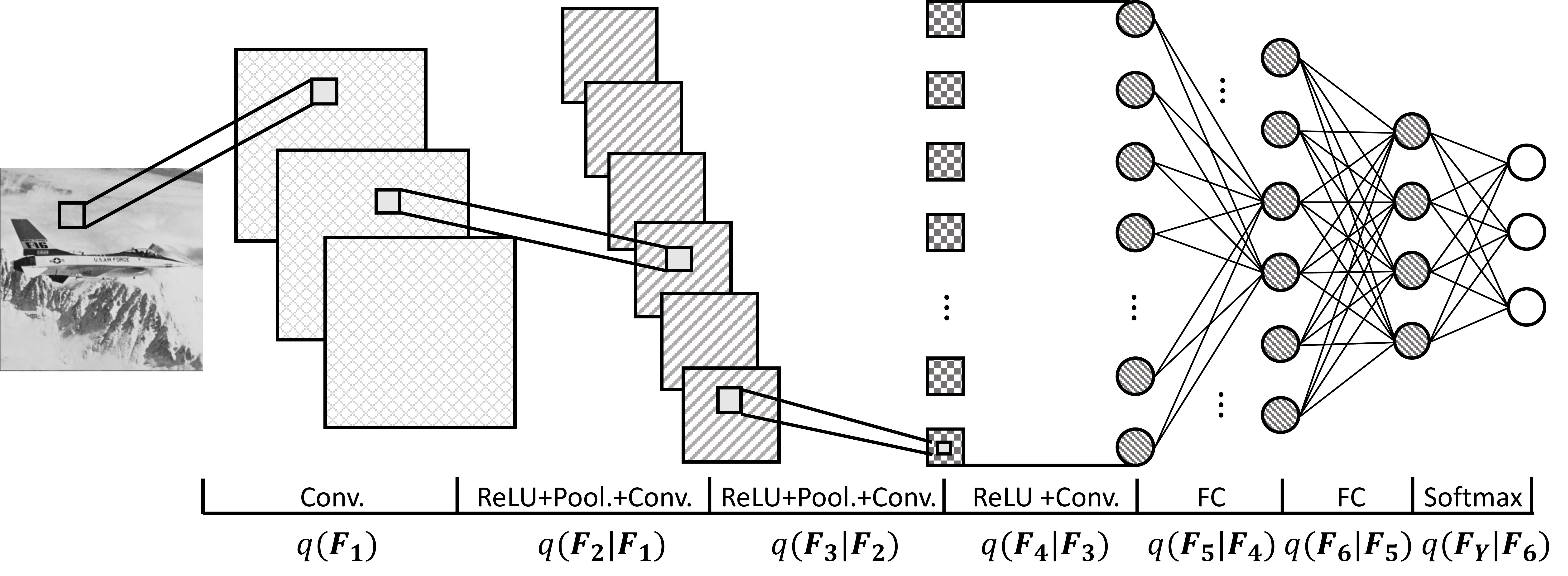}}
\end{minipage}
\hfill
\begin{minipage}[b]{0.28\linewidth}
\centerline{\includegraphics[scale=0.2]{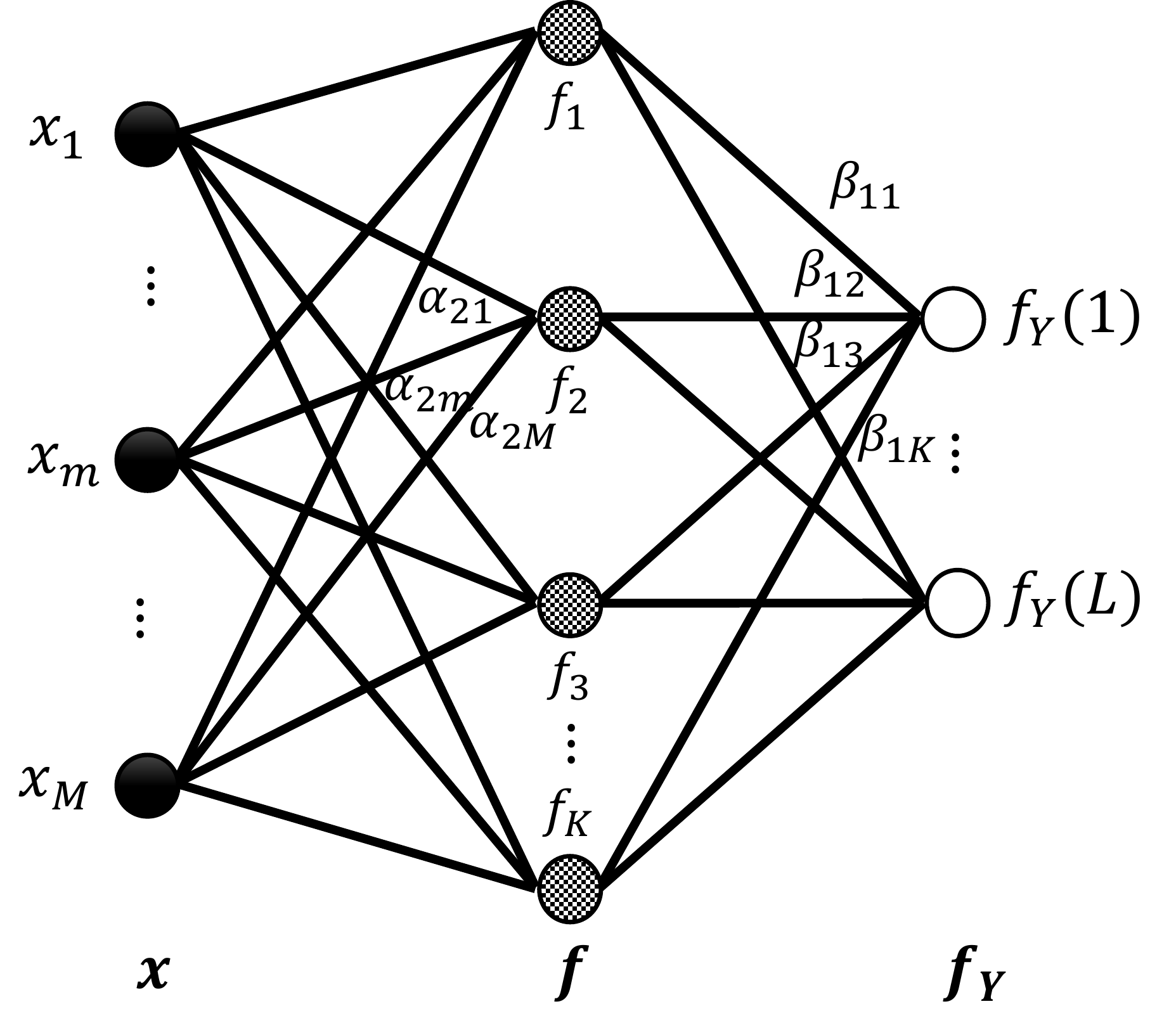}}
\end{minipage}
\caption{\small{ \textbf{(Left)}
The above DNN forms a Markov chain $\boldsymbol{F_1 \rightarrow \cdots \rightarrow F_6 \rightarrow F_Y}$, in which one or more hidden layers formulate a conditional Gibbs distribution. 
For example, the first fully connected (abbr. FC) layer describes the Gibbs distribution $q(\boldsymbol{F_5|F_4})$, and the second convolutional (abbr. Conv.) layer with a ReLU layer and a max pooling (abbr. Pool.) layer formulate $q(\boldsymbol{F_2|F_1})$.
Finally, the whole architecture of the DNN can be interpreted as a BHM $q(\boldsymbol{F_1}) \cdots q(\boldsymbol{F_4|F_3}) \cdots q(\boldsymbol{F_Y|F_6})$. 
\textbf{(Right)}
The shallow network $\{\boldsymbol{x; f; f_Y}\}$ is used for illustrating Proposition 2.
}}	
\label{fig_cnn_bhm}
\end{figure}

\textbf{Proposition 1:} \textit{The whole architecture of DNNs can be explained as a Bayesian hierarchical model.}

Since the input of the hidden layer $\boldsymbol{f_i}$ in the $\text{DNN} = \{\boldsymbol{x; f_1; ...; f_I; f_Y}\}$ is the output of its previous layer $\boldsymbol{f_{i-1}}$, 
we can derive that the $\text{DNN} = \{\boldsymbol{x; f_1; ...; f_I; f_Y}\}$ forms a Markov chain as
\begin{equation} 
\label{mrf_dnns}
\boldsymbol{F_1 \rightarrow \cdots \rightarrow F_I \rightarrow F_Y}\text{.}
\end{equation}
As a result, the distribution of the DNN can be formulated as
\begin{equation} 
\label{pdf_networks}
{\textstyle 
q(\boldsymbol{F_1; ...; F_I; F_Y}) = q(\boldsymbol{F_1}) \cdot ...q(\boldsymbol{F_{i+1}|F_{i}}) \cdot ... q(\boldsymbol{F_Y|F_{I}})\text{,}
}
\end{equation}
where $q(\boldsymbol{F_Y|F_{I}}) = \boldsymbol{f_Y}$ if the output layer is defined as softmax.
Notably, Proposition 2 shows that one or more hidden layers could be used to formulate a single conditional distribution in some cases. 
Since we can still derive a joint distribution in these cases, $q(\boldsymbol{F_1; ...; F_I; F_Y})$ is still used to indicate the joint distribution of DNNs for subsequent discussion.
An example is shown in Figure \ref{fig_cnn_bhm}.

The joint distribution $q(\boldsymbol{F_1; ...; F_I; F_Y})$ demonstrates that the DNN can be explained as a BHM with $I+1$ levels, in which the hidden layer $\boldsymbol{f_i}$ formulates a conditional distribution $q(\boldsymbol{F_{i}|F_{i-1}})$ to process the features in $\boldsymbol{f_{i-1}}$ and serves as a prior distribution for the higher level $\boldsymbol{f_{i+1}}$.
After establishing the probabilistic representation for the whole architecture of DNNs, we demonstrate that the hidden layers of DNNs can be explained as Gibbs distributions.
%The joint distribution $q(\boldsymbol{F_1; ...; F_I; F_Y})$ demonstrates that the DNN can be explained as a BHM with $I+1$ levels, in which the hidden layer $\boldsymbol{f_i}$ formulates a

\textbf{Proposition 2:} \textit{The hidden layers of a neural network formulate Gibbs distributions through defining the corresponding energy functions.} 

The intuition of Proposition 2 can be demonstrated by the shallow neural network $\{\boldsymbol{x; f; f_Y}\}$ shown in Figure \ref{fig_cnn_bhm}, in which the hidden layer $\boldsymbol{f}$ has $K$ neurons and the output layer $\boldsymbol{f_Y}$ is softmax with $L$ nodes. 
Therefore, we can formulate each output node as $f_{\boldsymbol{Y}}(l) = \frac {1}{Z}\text{exp}\{\sum_{k=1}^{K}\beta_{lk} \cdot g_k(\boldsymbol{x})\}$,
where $\boldsymbol{x} =\{x_1, \cdots, x_M \}$ is an input vector, ${\textstyle g_k(\boldsymbol{x})}$ formulates the $k$th neuron, and $\beta_{lk}$ denotes the weight of the edge between $f_{\boldsymbol{Y}}(l)$ and ${\textstyle g_k(\boldsymbol{x})}$.
The partition function is ${\scriptstyle Z = \sum_{\boldsymbol{x}}\sum_{l=1}^{L}\text{exp}\{\sum_{k=1}^{K}\beta_{lk} \cdot g_k(\boldsymbol{x})\}}$.

Previous works prove that $\boldsymbol{f_Y} = \{f_{\boldsymbol{Y}}(l)\}_{l=1}^{L}$ is equivalent to a discrete Gibbs distribution \cite{Gibbs_softmax}.
Specifically, $\boldsymbol{f_Y}$ assumes that there are $L$ configurations $ \{f_{\boldsymbol{Y}}(1), ..., f_{\boldsymbol{Y}}(L)\}$ of $\boldsymbol{x}$, and the energy of each configuration is expressed as ${\scriptstyle E_{\boldsymbol{Y}}(l) = -\sum_{k=1}^{K}\beta_{lk} \cdot g_k(\boldsymbol{x})}$.
Since $E_{\boldsymbol{Y}}(l)$ is a linear combination of all neurons $\{g_k(\boldsymbol{x})\}_{k=1}^{K}$, we can reformulate $f_{\boldsymbol{Y}}(l)$ as the Product of Expert (PoE) model \cite{CD}
\begin{equation} 
\label{snn_f} 
{\textstyle
%{\scriptstyle
f_{\boldsymbol{Y}}(l) = \frac{1}{Z'}\prod_{k=1}^{K}\{\frac {1}{Z_{\boldsymbol{F}}}\text{exp}[g_k(\boldsymbol{x})]\}^{\beta_{lk}} = \frac{1}{Z'}\prod_{k=1}^{K}\{p(F_{k})\}^{\beta_{lk}}
}\text{,}
\end{equation}
where ${\scriptstyle Z' = Z/\prod_{k=1}^{K}{Z_{\boldsymbol{F}}}^{\beta_{lk}}}$ and ${\scriptstyle Z_{\boldsymbol{F}} = \sum_{\boldsymbol{x}}\sum_{k=1}^{K}\text{exp}\{g_k(\boldsymbol{x})\}}$.

It is noteworthy that all experts $p(\boldsymbol{F}) = \{p(F_{k})\}_{k=1}^K$ are Gibbs distributions expressed as
\begin{equation} 
\label{snn_f1} 
{\textstyle
p(F_{k}) = \frac {1}{Z_{\boldsymbol{F}}}\text{exp}\{g_k(\boldsymbol{x})\}\text{,}
}
\end{equation}
where the energy function is equivalent to the negative of the $k$th neuron, i.e., $E_{\boldsymbol{F}}(k) = -g_k(\boldsymbol{x})$.
In other words, the energy function of $p(\boldsymbol{F})$ is entirely dependent on all the neurons in $\boldsymbol{f}$, namely the functionality of the hidden layer $\boldsymbol{f}$.
Since an energy function is a sufficient statistics of a Gibbs distribution \cite{it_book}, we can conclude that arbitrary hidden layers can be formulated as Gibbs distributions by defining the corresponding energy functions based on the functionality of the hidden layers.
%The detailed proof of Proposition 2 for DNNs is presented in the supplement A.

A straightforward example is DBM \cite{deep-BM}, in which each hidden layer defines an energy function as $E_{\text{RBM}}(\boldsymbol{f}) = -(\boldsymbol{b}_{H}^{T}\boldsymbol{f} + \boldsymbol{x}^{T}W\boldsymbol{f} + \boldsymbol{b}_{V}^{T}\boldsymbol{x})$, thereby formulating a special Gibbs distribution, namely Restricted Boltzmann Machine (RBM), 
\begin{equation} 
\label{bm} 
{\textstyle
p_{\text{RBM}}(\boldsymbol{F}) = \frac {1}{Z_{\boldsymbol{F}}}\text{exp}\{\boldsymbol{b}_{H}^{T}\boldsymbol{f} + \boldsymbol{x}^{T}\boldsymbol{Wf} + \boldsymbol{b}_{V}^{T}\boldsymbol{x}\}\text{,}
}
\end{equation}
where $\boldsymbol{b}_{H}$ and $\boldsymbol{b}_{V}$ are vectors of weights for the hidden nodes $\boldsymbol{f}$ and  the input vector $\boldsymbol{x}$, respectively. $\boldsymbol{W}$ is the matrix of connection weights. 
The partition function is $Z_{RBM} = \sum_{\boldsymbol{x, f}}\text{exp}\{-E_{\text{RBM}}(\boldsymbol{x})\}$.

In some cases, we should use multiple hidden layers to formulate a single Gibbs distribution. 
For example, a convolutional layer with non-linear layers have been proved to formulate the MRF model in the Convolutional Neural Networks (CNNs) \cite{xinjie2, CRF_RNN}.
Since MRF is a special Gibbs distribution \cite{Geman, MRF_IA}, we can conclude that a convolutional layer with non-linear layers formulate the energy function as $\boldsymbol{E}_{\text{Conv}}(\boldsymbol{x}) = -\sum_{k=1}^Kf_{k}[f_k^{NL}(\boldsymbol{x})]$ and define a Gibbs distribution expressed as
\begin{equation} 
\label{cnn_mrf} 
{\textstyle
p_{\text{Conv}}(\boldsymbol{F}) = \frac {1}{Z_{\boldsymbol{F}}}\text{exp}\{\sum_{k=1}^Kf_{k}[f_k^{NL}(\boldsymbol{x})]\}\text{,}
}
\end{equation}
where $\boldsymbol{x} \in \boldsymbol{\mathcal{D}}$ is a high-dimensional input, $f_k(\cdot)$ is a convolutional filter, and $f^{NL}(\cdot)$ denote non-linear layer(s), such as ReLU. 
The partition function is $Z_{\text{Conv}} = \sum_{\boldsymbol{x}}\text{exp}\{-\boldsymbol{E}_{\text{Conv}}(\boldsymbol{x})\}$.

It needs to be emphasized that hidden layers only formulate the corresponding energy functions, rather than directly formulating Gibbs distributions.
We can conclude that hidden layers formulate Gibbs distributions because an energy function is a sufficient statistics of a Gibbs distribution \cite{it_book}.

Overall, the above two propositions provide an explicitly probabilistic representation of DNNs in three aspects: 
(i) neurons define the energy of a Gibbs distribution;
(ii) the hidden layers of DNNs formulate Gibbs distributions;
and (iii) the whole architecture of DNNs can be interpreted as a BHM.
Based on the probabilistic representation, we provide insights into two fundamental properties of deep learning, i.e., hierarchy and generalization, in the next section.

\section{Insights into deep learning}

\subsection{Hierarchy}
\label{hierarchy}

Based on Proposition 1, we can explicitly formulate the hierarchy property of deep learning.  
More specifically, the $\text{DNN} = \{\boldsymbol{x; f_1; ...; f_I; f_Y}\}$ describes a BHM as $q(\boldsymbol{F_1; ...; F_I; F_Y})$ to simulate the joint distribution $p_{\boldsymbol{\theta}}(\boldsymbol{X}, \boldsymbol{Y})$ given $\boldsymbol{\mathcal{D}}$, which can be expressed as
\begin{equation} 
\label{bayesian_networks}
{\textstyle 
q(\boldsymbol{F_1; ...; F_I; F_Y}) = \underbrace{{\textstyle q(\boldsymbol{F_1})  \cdots q(\boldsymbol{F_{i-1}|F_{i-2}})}}_{{\scriptstyle \text{prior}}} \underbrace{{\textstyle \cdot q(\boldsymbol{F_i|F_{i-1}}) \cdots q(\boldsymbol{F_Y|F_{I}})}}_{{\scriptstyle \text{likelihood}}}\text{.}
}
\end{equation}
This equation indicates that the DNN uses some hidden layers (i.e., $\boldsymbol{f_1} \cdots \boldsymbol{f_{i-1}}$) to learn a prior distribution $q(\boldsymbol{X})$ and the other layers (i.e., $\boldsymbol{f_i} \cdots \boldsymbol{f_Y}$) to learn a likelihood distribution $q(\boldsymbol{Y|X})$.
For simplicity, DNNs formulates a joint distribution $q(\boldsymbol{X, Y}) = q(\boldsymbol{Y|X})q(\boldsymbol{X})$ to model $p_{\boldsymbol{\theta}}(\boldsymbol{X}, \boldsymbol{Y})$.

Compared to traditional Bayesian models, there are two characteristics of DNNs. 
First, there is no clear boundary to separate DNNs into two parts, i.e., $q(\boldsymbol{X})$ and $q(\boldsymbol{Y|X})$, because the architecture of DNNs is much more complex than an ordinary BHM \cite{BHM}.
Second, unlike the naive Bayes classifier independently inferring the parameters of $q(\boldsymbol{X})$ and $q(\boldsymbol{Y|X})$ \cite{dis_vs_gen} from $\boldsymbol{\mathcal{D}}$, the learning algorithm of DNNs, e.g., backpropagation \cite{backpropagation}, infers the parameters of $q(\boldsymbol{X})$ based on that of $q(\boldsymbol{Y|X})$. 
These characteristics lead to both pros and cons of DNNs.
On the one hand, they enable DNNs to freely learn various features from $\boldsymbol{\mathcal{D}}$ to formulate $q(\boldsymbol{X, Y})$.
On the other hand, they result in some inherent problems of DNNs, such as overfitting, which is discussed in the next section.

\subsection{Generalization}

%In contrast to previous work claiming that DNNs perform implicit {regularization} by SGD \cite{Soatto1, gd_vr, generalization_regularization1, generalization_regularization},
Based on the hierarchy property of DNNs, we can demonstrate DNNs having an explicit {regularization} because the Bayesian theory indicates that a prior distribution corresponds to the regularization \cite{bayesian_regularization}.
This novel insight explains why an over-parametrized DNN still can achieve great generalization performance \cite{regularization5}. 
More specifically, though an over-parametrized DNN indicates that it has a much complex $q(\boldsymbol{Y|X})$, it simultaneously implies that the DNN can use many hidden units to formulate a powerful $q(\boldsymbol{X})$ to regularize $q(\boldsymbol{Y|X})$, thereby guaranteeing the generalization performance.

%The next question need to answer is why DNNs are still overfitting in some cases if DNNs have an explicit regularization?
%This is because the special learning algorithm, e.g., the backpropagation \cite{backpropagation}, decreases the generalization ability of DNNs. 
Moreover, we demonstrate that the learning algorithm, e.g., the backpropagation \cite{backpropagation}, is the reason for decreasing the generalization ability of DNNs.
Given a $\text{DNN} = \{\boldsymbol{x; f_1; ...; f_I; f_Y}\}$, we can formulate a BHM as $q(\boldsymbol{F_1; ...; F_I; F_Y})$.
From the perspective of variational inference \cite{gd_vr, SVI}, the backpropagation aims to find an optimal distribution $q^*(\boldsymbol{F_1; ...; F_I; F_Y})$ that minimizes the KL divergence to the truly posterior distribution $p(\boldsymbol{F_1; ...; F_I; F_Y}|\boldsymbol{\mathcal{D}})$. 
\begin{equation} 
\label{VI} 
{\textstyle q^{*}(\boldsymbol{F_1; ...; F_I; F_Y}) = \argmin_{q \in \mathcal{Q}} \text{KL}[p(\boldsymbol{F_1; ...; F_I; F_Y}|\boldsymbol{\mathcal{D}})||q(\boldsymbol{F_1; ...; F_I; F_Y})]} \\
\end{equation}
Ideally, this optimization problem is expected to be solved by iteratively optimizing each random variable $\boldsymbol{F_i}$ while holding other random variables $\boldsymbol{F_{-i}} = \{\boldsymbol{F_1; ...;F_{i-1}; F_{i+1}; ...; F_I; F_{Y}}\}$ fixed. 
\begin{equation} 
\label{VI_single_layer} 
{\textstyle q^{*}(\boldsymbol{F_i |F_{-i}}) = \argmin_{q \in \mathcal{Q}}\text{KL}[p(\boldsymbol{F_{i}|F_{-i}; \boldsymbol{\mathcal{D}}})||q(\boldsymbol{F_i|F_{-i}})]}
\end{equation}
However, we cannot derive $q^{*}(\boldsymbol{F_1; ...; F_I; F_Y})$ in practice because $p(\boldsymbol{F_{i}|F_{-i}; \mathcal{D}})$ is intractable.

To design a feasible learning algorithm for DNNs, the loss function is alternatively relaxed to
\begin{equation} 
\label{VI_output_layer} 
{\textstyle q^{*}(\boldsymbol{F_Y|F_1; ...; F_I}) = \argmin_{q \in \mathcal{Q}}\text{KL}[p(\boldsymbol{F_{Y}}|\{\boldsymbol{y}_n\}_{n=1}^N)||q(\boldsymbol{F_Y|F_1; ...; F_I})]}
\end{equation}
because $p(\boldsymbol{F_{Y}}|\boldsymbol{F_1; ...; F_I; \mathcal{D}}) = p(\boldsymbol{F_{Y}}|\{\boldsymbol{y}_n\}_{n=1}^N)$ is known to us.
Nevertheless, the cost for this relaxation is that we cannot precisely infer $p(\boldsymbol{F_1; ...; F_I; F_Y}|\boldsymbol{\mathcal{D}})$.
For simplicity, the truly posterior distribution $p(\boldsymbol{F_1; ...; F_I; F_Y}|\boldsymbol{\mathcal{D}})$ can be expressed as $p_{\boldsymbol{\theta}}(\boldsymbol{Y},\boldsymbol{X})$, thus the loss function for DNNs should be formulated as $\text{KL}[p_{\boldsymbol{\theta}}(\boldsymbol{Y},\boldsymbol{X})||q(\boldsymbol{Y}, \boldsymbol{X})] = \text{KL}[p(\boldsymbol{Y}|\boldsymbol{X})||q(\boldsymbol{Y}|\boldsymbol{X})] + \text{KL}[p(\boldsymbol{X})||q(\boldsymbol{X})]$.
However, the relaxed loss function merely corresponds to $\text{KL}[p(\boldsymbol{Y}|\boldsymbol{X})||q(\boldsymbol{Y}|\boldsymbol{X})]$, which implies that it cannot guarantee the learned DNNs satisfying the generalization property of DNNs.

It is noteworthy that the conjugacy property, namely that both $q(\boldsymbol{Y}|\boldsymbol{X})$ and $q(\boldsymbol{Y}, \boldsymbol{X})$ are Gibbs distributions derived from the same DNN, and the backpropagation inferring parameters in the backward direction enable us to infer $q(\boldsymbol{X})$ via $q(\boldsymbol{Y}|\boldsymbol{X})$ based on the relaxed loss function.
However, the primary goal of the learned $q(\boldsymbol{X})$ is to derive the $q(\boldsymbol{Y|X})$ that is close to $p(\boldsymbol{Y}|\boldsymbol{X})$  but not to precisely model the truly prior distribution $p(\boldsymbol{X})$.
% though the backpropagation inferring parameters in the backward direction enables DNNs to learn $q(\boldsymbol{X})$ via $q(\boldsymbol{Y|X})$.
% the .

\section{Experiments}
\label{expe}

%In this section, %we carry out experiments to validate the proposed probabilistic representation and the novel insights into the hierarchy and generalization properties.
In this section, we first demonstrate the proposed probabilistic representation and the hierarchy property based on a simple but comprehensive CNN on a synthetic dataset.
Subsequently, we validate the proposed insights into the generalization property and clarify two notable empirical empirical phenomena of deep learning that cannot be explained by the traditional theories of generalization. 
%show that directly learning $q(\boldsymbol{X})$ from $\boldsymbol{\mathcal{D}}$ is an effective \textit{regularization} for improving \textit{generalization} performance for DNNs based on the CIFAR-10 classification task.
%All simulation codes are available in the supplement. 
%We also report some experiments in the supplement due to the space restriction.

\subsection{The proposed probabilistic representation}

\begin{figure}[t]
\centering
\includegraphics[scale=0.4]{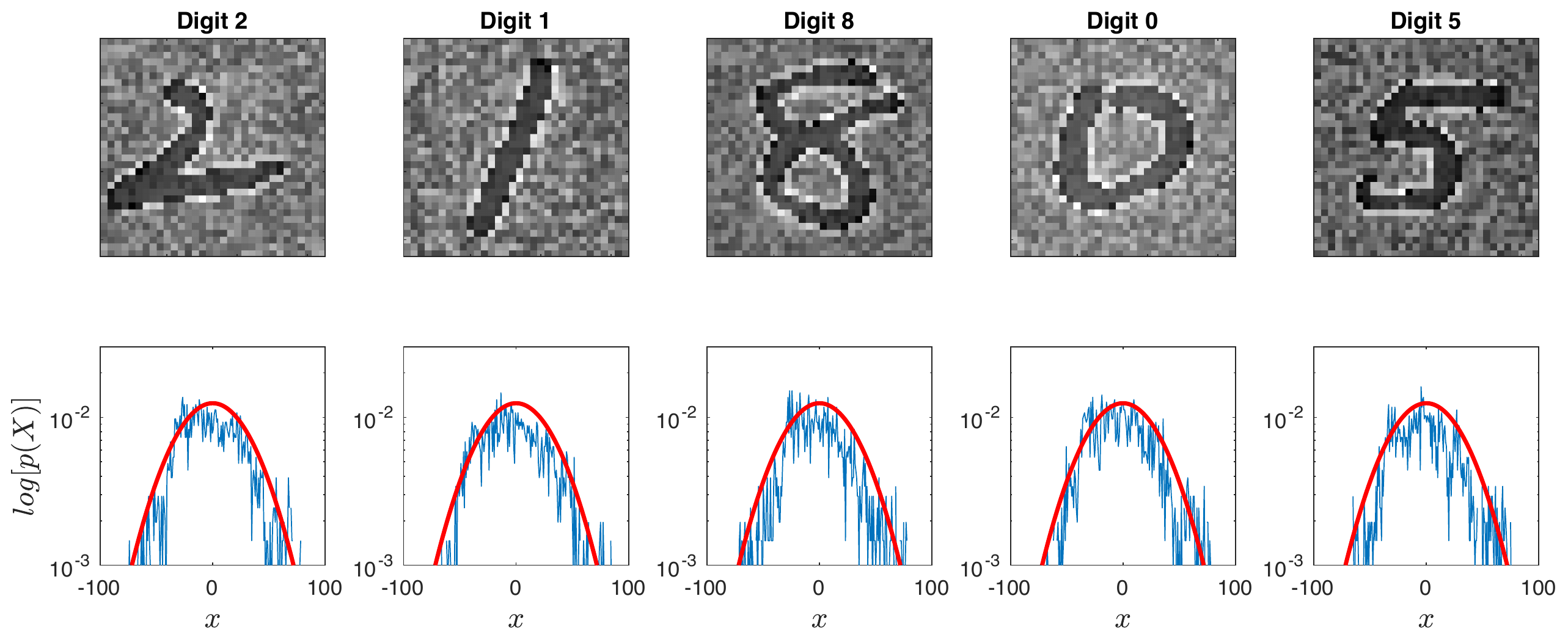}
\caption{\small{
The first row shows five synthetic images of handwritten digits, the second row shows their respective histograms, and the red curve indicates the Gaussian distribution $\mathcal{N}(0, 1024)$.
}}
\label{fig_synthetic_gaussian}
\end{figure}

Since the distributions of most benchmark datasets are unknown, it is impossible to use them to demonstrate the proposed probabilistic representation.
Alternatively, we generate a synthetic dataset obeying the Gaussian distribution $\mathcal{N}(0, 1024)$ based on the NIST dataset of handwritten digits \footnote{\url{https://www.nist.gov/srd/nist-special-database-19}}.
The synthetic dataset consists of 20,000 $32 \times 32$ grayscale images in 10 classes (digits from 0 to 9). 
All grayscale images are sampled from the Gaussian distribution $\mathcal{N}(0, 1024)$.
Each class has 1,000 training images and 1,000 testing images. 
Figure \ref{fig_synthetic_gaussian} shows five synthetic images and their perspective histograms.
The method for generating the synthetic dataset is reported in the supplement A.

We choose CNN1 from Table \ref{cnns_synthetic} to classify the synthetic dataset.
Based on the proposed probabilistic representation, we can identify the functionality of each hidden layer as follows.
Above all, $\boldsymbol{f_1}$ should model the truly prior distribution i.e., $q(\boldsymbol{F_1}) \approx p(\boldsymbol{X})$, because the max pooling layer $\boldsymbol{f_2}$ compresses too much information of $\boldsymbol{x}$ for dimension reduction.
The subsequent hidden layers formulate $q(\boldsymbol{F_2|F_1})$ and $q(\boldsymbol{F_Y|F_2})$, thereby modeling the likelihood distribution $p(\boldsymbol{Y|X})$.
In summary, the whole architecture of CNN1 formulates a BHM as $q(\boldsymbol{F_1; F_2; F_Y}) = q(\boldsymbol{F_1})q(\boldsymbol{F_2|F_1})q(\boldsymbol{F_Y|F_2})$.

\begin{table}[!b]
\caption{The architectures of CNNs for experiments}
\label{cnns_synthetic}
\vskip 0.15in
\begin{center}
\begin{small}
%\begin{sc}
\begin{threeparttable}
\begin{tabular}{ccccc}
\toprule
R.V. & Layer & Description & CNN1 & CNN2 \\
\midrule
$\boldsymbol{X}$&$\boldsymbol{x}$					& Input 				& $32 \times 32 \times 1$ & $32 \times 32 \times 1$ \\
\hline
$\boldsymbol{F_1}$ &$\boldsymbol{f_1}$				& Conv ($3 \times 3$)	& $30 \times 30 \times {20}$ & $30 \times 30 \times {20}$ \\
\hline 
\multirow{2}*{$\boldsymbol{F_2}$}&$\boldsymbol{f_2}$    	& Maxpool + ReLU		& $15 \times 15 \times {20}$ & $15 \times 15 \times 20$ \\
&$\boldsymbol{f_3}$    							& Conv ($5 \times 5$)	& $11 \times 11 \times \boldsymbol{60}$ & $11 \times 11 \times \boldsymbol{36}$ \\
\hline 
\multirow{3}*{$\boldsymbol{F_Y}$}&$\boldsymbol{f_4}$    	& Maxpool + ReLU		& $5 \times 5 \times {60}$ & $5 \times 5 \times 36$ \\
&$\boldsymbol{f_5}$      							& Fully connected		& $1 \times 1 \times 10$  & $1 \times 1 \times 10$ \\
&$\boldsymbol{f_Y}$   							& Output(softmax)		& $1 \times 1 \times 10$  & $1 \times 1 \times 10$ \\
\bottomrule
\end{tabular}
\begin{tablenotes}
            \item R.V. is the random variable of the hidden layer(s), and the only difference between CNN1 and CNN2 is the number of convolutional filters in $\boldsymbol{f_3}$.
   \end{tablenotes}
%\end{sc}
\end{threeparttable}
\end{small}
\end{center}
\vskip -0.1in
\end{table}

Since $\boldsymbol{f_1}$ and $\boldsymbol{f_3}$ are convolutional layers, we can formulate $q(\boldsymbol{F_1})$ and $q(\boldsymbol{F_2|F_1})$ as 
%based on Equation (\ref{cnn_mrf}). 
\begin{equation} 
\label{cnn_mrf_prior} 
%{\scriptstyle
{\textstyle
q(\boldsymbol{F_1}) = \frac {1}{Z_{\boldsymbol{F_1}}}\text{exp}\{\sum_{k=1}^{20}f_{k}(\boldsymbol{x})\}\text{ and }q(\boldsymbol{F_2|F_1}) = \frac {1}{Z_{\boldsymbol{F_2}}}\text{exp}\{\sum_{k=1}^{60}f'_{k}[f^{NL}(\boldsymbol{f_1})]\}}\text{,}
\end{equation}
where $f_k$ is a $3 \times 3$ convolutional filter in $\boldsymbol{f_1}$, $f'_k$ is a $5 \times 5$ convolutional filter in $\boldsymbol{f_3}$, and $f^{NL}$ indicate the max pooling and ReLU operators in $\boldsymbol{f_2}$ and $\boldsymbol{f_4}$. 
In addition, ${\scriptstyle \boldsymbol{E}_{\boldsymbol{F_1}}(\boldsymbol{x}) = -\sum_{k=1}^{20}f_{k}(\boldsymbol{x})}$, ${\scriptstyle \boldsymbol{E}_{\boldsymbol{F_2}}(\boldsymbol{f_1}) = -\sum_{k=1}^{60}f'_{k}[f^{NL}(\boldsymbol{f_1})]}$, ${\scriptstyle Z_{\boldsymbol{F_1}} = \sum_{\boldsymbol{x}}\text{exp}\{-\boldsymbol{E}_{\boldsymbol{F_1}}(\boldsymbol{x})\}}$, and ${\scriptstyle Z_{\boldsymbol{F_2}} = \sum_{\boldsymbol{f_1}}\text{exp}\{-\boldsymbol{E}_{\boldsymbol{F_2}}(\boldsymbol{f_1})\}}$.
Since the output layer $\boldsymbol{f_Y}$ is defined as softmax, the output nodes $\{q(\boldsymbol{F_Y}(l)|\boldsymbol{F_2})\}_{l=1}^{10}$ can be expressed as
\begin{equation} 
%{\scriptstyle
{\textstyle
q(\boldsymbol{F_Y}(l)|\boldsymbol{F_2}) = \frac{1}{Z_{\boldsymbol{F_Y}}}\text{exp}\{\sum_{k=1}^{60}\beta_{lk} \cdot f^{''}_{k}[f^{NL}(\boldsymbol{f_3})]\}\text{,}
}
\end{equation}
where $f^{''}_{k}$ is the linear filter in $\boldsymbol{f_5}$, and $\beta_{lk}$ is the weight of the edge between $\boldsymbol{f_5}$ and $\boldsymbol{f_Y}$.

Though we obtain the formulas of $q(\boldsymbol{F_1})$ and $q(\boldsymbol{F_2|F_1})$, it is hard to calculate $q(\boldsymbol{F_1})$ and $q(\boldsymbol{F_2|F_1})$ because $Z_{\boldsymbol{F_1}}$ and $Z_{\boldsymbol{F_2}}$ are intractable for the high dimensional datasets $\boldsymbol{x}$ and $\boldsymbol{f_1}$.
Alternatively, we use the histograms of $\boldsymbol{E}_{\boldsymbol{F_1}}$ and $\boldsymbol{E}_{\boldsymbol{F_2}}$ to estimate $q(\boldsymbol{F_1})$ and $q(\boldsymbol{F_2|F_1})$, respectively, because an energy function is a sufficient statistics of a Gibbs distribution \cite{it_book, statistical_img, GSM-stat}.

\begin{figure}[t]
\centering
\includegraphics[scale=0.46]{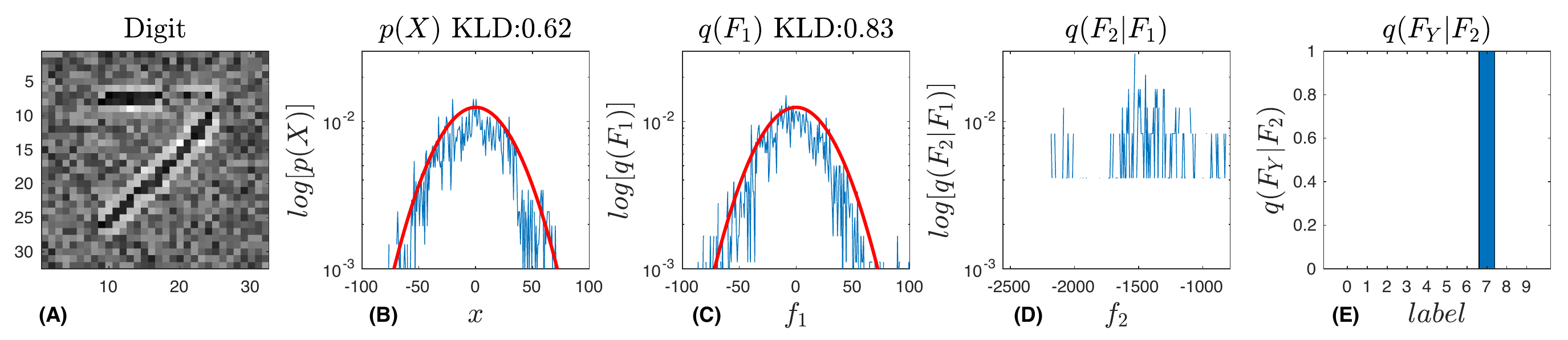}
\caption{\small{
The distribution of the hidden layers in CNN1. The red curve indicates the truly prior distribution $p(\boldsymbol{X}) = \mathcal{N}(0, 1024)$.
The blue curves are different histograms.
(A) the synthetic image $\boldsymbol{x}$ is the input of CNN1.
(B) the histogram of $\boldsymbol{x}$ and $\text{KL}[p(\boldsymbol{X})||p(\boldsymbol{x})] = 0.62$.
(C) the histogram of $\boldsymbol{E}_{\boldsymbol{F_1}}$ for estimating $q(\boldsymbol{F_1})$ and $\text{KL}[p(\boldsymbol{X})||q(\boldsymbol{F_1})] = 0.83$.
(D) the histogram of $\boldsymbol{E}_{\boldsymbol{F_2}}$ for estimating $q(\boldsymbol{F_2})$.
(E) the output is $q(\boldsymbol{F_Y|F_2})$.
}}
\label{fig_cnn_sim1}
\end{figure}

After CNN1 is well trained (i.e., the training error becomes zero), we randomly choose a testing image $\boldsymbol{x}$ as the input of CNN1 for deriving $q(\boldsymbol{F_1})$, $q(\boldsymbol{F_2|F_1})$, and $q(\boldsymbol{F_Y|F_2})$. 
Since $\boldsymbol{x}$ is a sample generated from $p(\boldsymbol{X})$, i.e., $\boldsymbol{x} \sim p(\boldsymbol{X}) = \mathcal{N}(0, 1024)$, we can exam the proposed probabilistic representation through calculating the distance between $q(\boldsymbol{F_1})$ and $p(\boldsymbol{X})$, i..e, $\text{KL}[p(\boldsymbol{X})||q(\boldsymbol{F_1})]$, to check if $\boldsymbol{f_1}$ precisely models $p(\boldsymbol{X})$.
All distributions are shown in Figure \ref{fig_cnn_sim1}.
We see that $q(\boldsymbol{F_1})$ is very close to $p(\boldsymbol{X})$ ($\text{KL}(p(\boldsymbol{X})||q(\boldsymbol{F_1})) = 0.83$) and $q(\boldsymbol{F_Y|F_2})$ outputs correct classification probability.

This experiment validates the proposed probabilistic representation in two aspects.
First, since we can theoretically prove CNN1 formulating a joint distribution $q(\boldsymbol{F_1; F_2; F_Y})$ and empirically show $q(\boldsymbol{F_1})$ modeling the prior distribution $p(\boldsymbol{X})$, we can conclude that $\text{CNN1} = \{\boldsymbol{x; f_1; ...; f_5; f_Y}\}$ formulates a BHM as $q(\boldsymbol{F_1; F_2; F_Y})$, thereby explaining the {hierarchy} property of DNNs.
Second, the hidden layers of CNN1, e.g., $\boldsymbol{f_1}$, formulate Gibbs distributions by defining the corresponding energy function.
Moreover, it shows that DNNs have an explicit regularization by learning a prior distribution and preliminarily validates the novel insight into the generalization property of DNNs.

\subsection{Generalization}

\begin{figure}[!b]
\centering
\begin{minipage}[b]{0.49\linewidth}
\centerline{\includegraphics[scale=0.35]{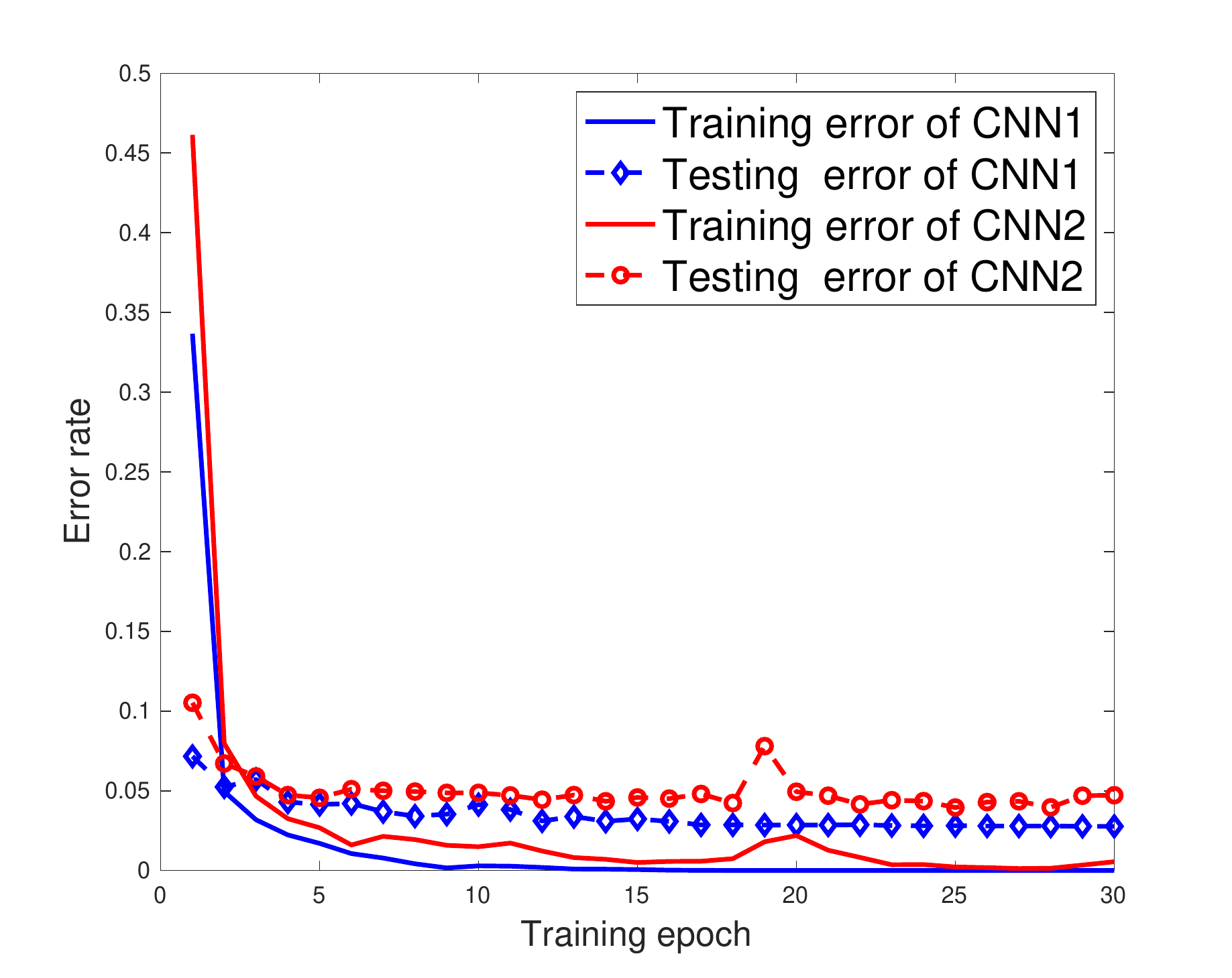}}
\end{minipage}
\hfill
\begin{minipage}[b]{0.49\linewidth}
\centerline{\includegraphics[scale=0.35]{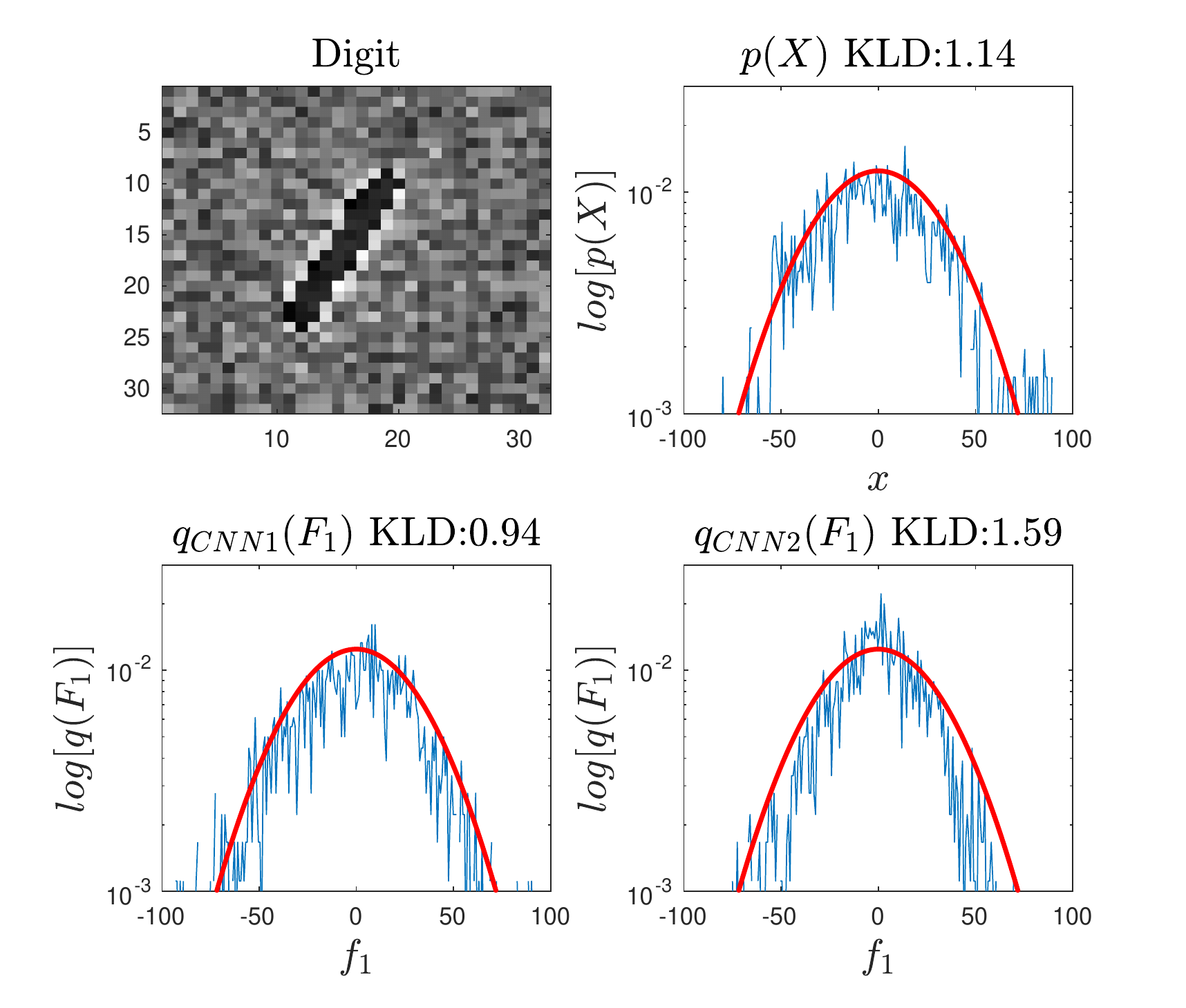}}
\end{minipage}
\caption{\small{ \textbf{(Left)} 
The generalization performance of CNN1 and CNN2 is represented by the testing error.
\textbf{(Right)}
A synthetic image and the histograms of $p(\boldsymbol{X})$, $q_\text{CNN1}(\boldsymbol{F_1})$, and $q_\text{CNN2}(\boldsymbol{F_1})$ given the synthetic image. 
}}	
\label{fig_cnn_generalization}
\end{figure}

\subsubsection{Analyzing the generalization property of over-parametrized DNNs}

%The first experiment shows that DNNs have an explicit regularization by learning a prior distribution. 
%In the context of Bayesian theory \cite{bayesian_regularization}, a better prior distribution should achieve better generalization performance given the same likelihood distribution.
We use CNN1 and CNN2 to further validate the insights into the generalization property by comparing their performances on the synthetic dataset.
Notably, both CNN1 and CNN2 are over-parametrized because a synthetic image has 1024 pixels but they have 1680 and 1330 parameters, respectively.
%In addition, the only difference between CNN1 and CNN2 is that the former has much more convolutional filters than the later in $\boldsymbol{f_3}$.

Based on the proposed probabilistic representation, Table \ref{cnns_synthetic} indicates that CNN1 and CNN2 have the same convolutional layer $\boldsymbol{f_1}$ to formulate their respective prior distributions, i.e., $q_\text{CNN1}(\boldsymbol{F_1})$ and $q_\text{CNN2}(\boldsymbol{F_1})$. 
Meanwhile, CNN1 formulates a much more complex likelihood distribution than CNN2 because the former has much more convolutional filters than the later in $\boldsymbol{f_3}$.
Intuitively, given the same complexity of the prior distributions, a more complex likelihood distribution is more prone to be overfitting, but Figure \ref{fig_cnn_generalization} shows that CNN1 has the better generalization performance than CNN2.
Also note that $\text{KL}[p(\boldsymbol{X})||q_\text{CNN1}(\boldsymbol{F_1})] = 0.94$ and $\text{KL}[p(\boldsymbol{X})||q_\text{CNN2}(\boldsymbol{F_1})] = 1.59$.
%the likelihood distribution of CNN1 should be much more complex than that of CNN2.
%Intuitively, more complex likelihood distribution is more prone to be overfitting, but Figure \ref{fig_cnn_generalization} shows that CNN1 achieves better generalization performance than CNN2.
%Also note that $\text{KL}(p(\boldsymbol{X})||q_\text{CNN1}(\boldsymbol{F_1})) = 0.94$ and $\text{KL}(p(\boldsymbol{X})||q_\text{CNN2}(\boldsymbol{F_1})) = 1.59$,
It means that CNN1 learns the better prior distribution, thus it can regularize the likelihood distribution of CNN1 better and guarantee its superiority over CNN2 even though they are over-parametrized networks.
%This experiment further validates DNNs having an explicit regularization.
%That explains why increasing the number of hidden units in over-parametrized networks can decrease the generalization error but not result in overfitting.

Moreover, this experiment shows that the learning algorithm limits the generalization ability of DNNs. 
Equation (\ref{cnn_mrf_prior}) indicates that $q_\text{CNN1}(\boldsymbol{F_1})$ and $q_\text{CNN2}(\boldsymbol{F_1})$ have the same formula.
Intuitively, given the same training dataset $\boldsymbol{\mathcal{D}}$, the learned $q_\text{CNN2}(\boldsymbol{F_1})$ should be as good as $q_\text{CNN1}(\boldsymbol{F_1})$.
However, Figure \ref{fig_cnn_generalization} shows that $q_\text{CNN2}(\boldsymbol{F_1})$ is worse than $q_\text{CNN1}(\boldsymbol{F_1})$ though both CNN1 and CNN2 achieve zero training error.
It implies that the relaxed loss function cannot guarantee the backpropagation accurately inferring the prior distribution.
Specifically, since the backpropagation infers the parameters of DNNs in the backward direction \cite{backpropagation}, it has to infer the prior distributions via the likelihood distributions rather than directly from $\boldsymbol{\mathcal{D}}$. 
As a result, the hidden layers corresponding to the likelihood distributions, especially $\boldsymbol{f_3}$, have great effect on inferring $q_\text{CNN1}(\boldsymbol{F_1})$ and $q_\text{CNN2}(\boldsymbol{F_1})$. 
In particular, Table \ref{cnns_synthetic} shows that CNN1 has 60 convolutional filters in $\boldsymbol{f_3}$ but CNN2 only has 36, thus the backpropagation cannot infer $q_\text{CNN2}(\boldsymbol{F_1})$ as accurate as $q_\text{CNN1}(\boldsymbol{F_1})$, thereby limiting the generalization ability of CNN2.
%Moreover, since the backpropagation only minimizes $\text{KL}[p(\boldsymbol{Y}|\boldsymbol{X})||q(\boldsymbol{Y}|\boldsymbol{X})]$.  
%Moreover, since the backpropagation only minimizes $\text{KL}[p(\boldsymbol{Y}|\boldsymbol{X})||q(\boldsymbol{Y}|\boldsymbol{X})]$, .  

\begin{figure}[t]
\centering
\includegraphics[scale=0.45]{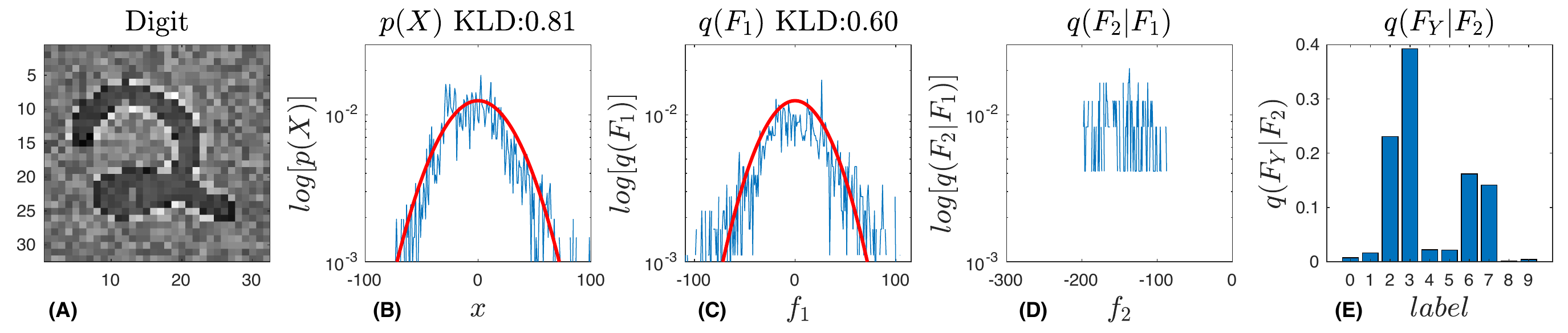}
\caption{\small{
The distribution of the hidden layers in CNN1 given a synthetic testing image with the random label 8. 
The notation is the same as Figure \ref{fig_cnn_sim1}, e.g., $\text{KL}[p(\boldsymbol{X})||p(\boldsymbol{x})] = 0.81$, and $\text{KL}[p(\boldsymbol{X})||q(\boldsymbol{F_1})] = 0.60$.
}}
\label{fig_distribution_random}
\end{figure}

\subsubsection{Analyzing the generalization property of DNNs on random labels}

\begin{figure}[htp]
\centering
\begin{minipage}[b]{0.39\linewidth}
\centerline{\includegraphics[scale=0.23]{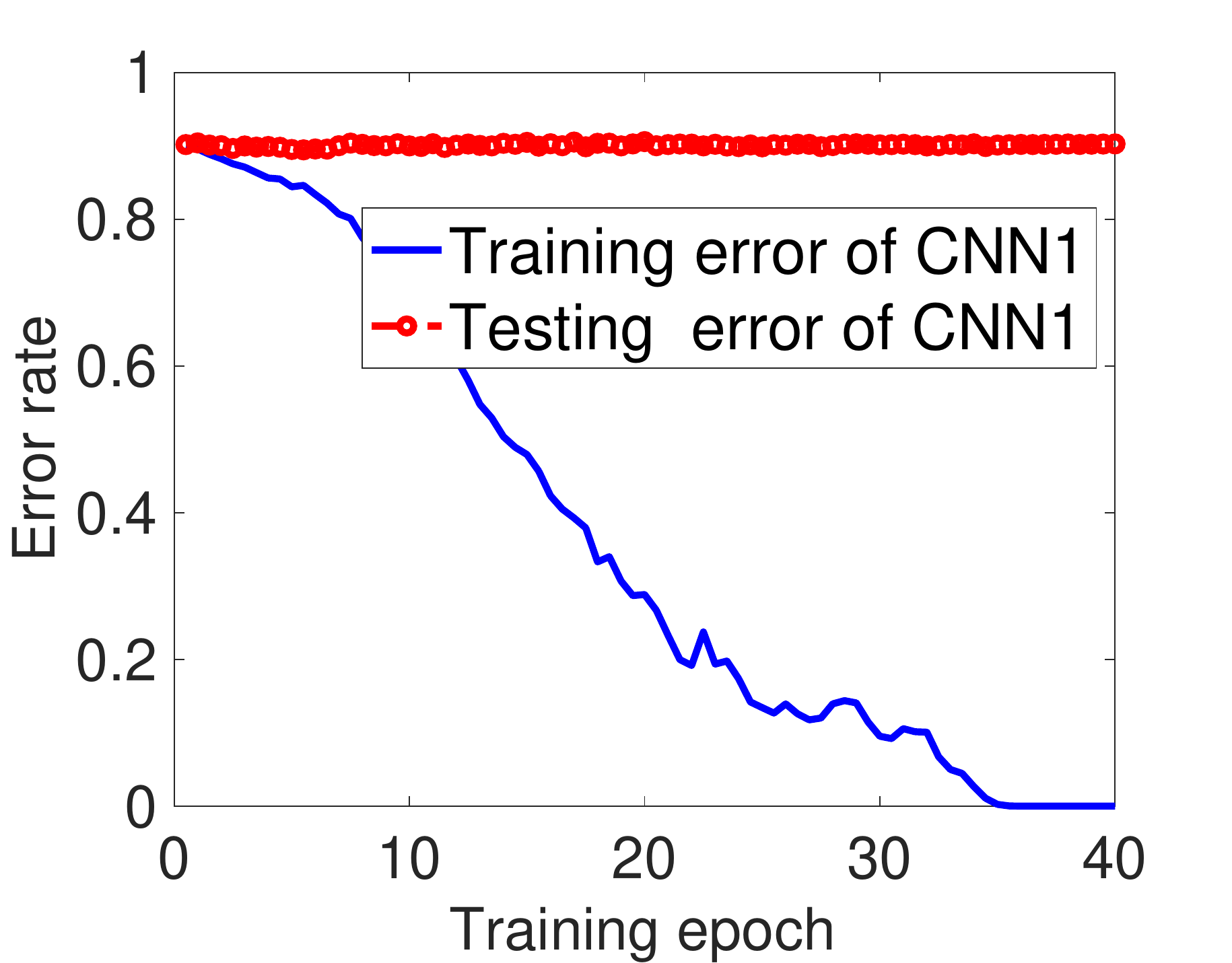}}
\end{minipage}
\begin{minipage}[b]{0.39\linewidth}
\centerline{\includegraphics[scale=0.25]{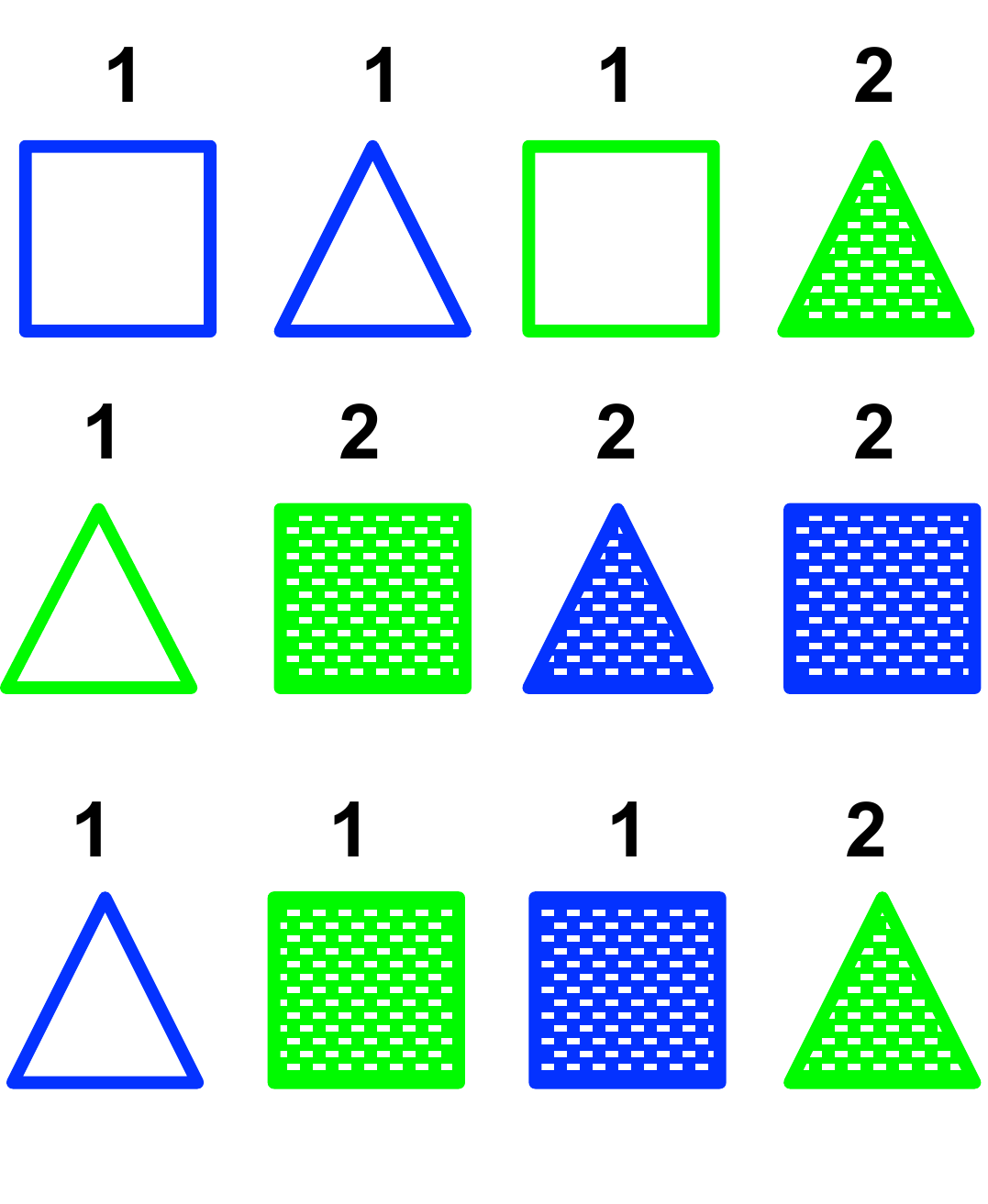}}
\end{minipage}
\caption{\small{ \textbf{(Left)} 
The performance of CNN1 for the synthetic dataset with random labels.
\textbf{(Right)}
The number above the object indicates a random label. The objects in the two upper rows indicate the training dataset, and the objects in the last row indicate the testing dataset.
}}	
\label{fig_cnn_random}
\end{figure}

Similar to the experiment presented in \cite{generalization_regularization}, we use CNN1 to classify the synthetic dataset with random labels. 
Figure \ref{fig_cnn_random} shows that CNN1 achieves zero training error but very high testing error.
We also visualize the distribution of each hidden layer in CNN1 given a testing image with a random label, and Figure \ref{fig_distribution_random}(C) shows that CNN1 still can learn an accurate prior distribution $q(\boldsymbol{F_1})$.
In this sense, CNN1 still achieves good generalization performance for this experiment.

We use a simple example to demonstrate that it is impossible for a DNN to classify random labels because it can only model two dependent random variables, i.e., $q(\boldsymbol{X}, \boldsymbol{Y}) = q(\boldsymbol{Y|X})q(\boldsymbol{X})$, but random label implies $p_{\boldsymbol{\theta}}(\boldsymbol{X}, \boldsymbol{Y}) = p(\boldsymbol{Y})p(\boldsymbol{X})$. 
In this example, we assume that humans can only distinguish the shape of the object (triangle or square), but the color feature (blue or green) and whether the object is filled (full or empty) are hidden features for humans.
Since we can only detect the shape feature, there are only two label values (1 and 2). 
Labels are randomly assigned to 12 objects (8 for training and 4 for testing) in Figure \ref{fig_cnn_random}.  
It is known that DNNs can detect many features that are imperceptible for humans, thus we assume that DNNs can detect all the features.
Given the training objects with random labels, DNNs extract all three features as prior knowledge, i.e., $q(\boldsymbol{X}) = q(\text{feature})$, and find that objects can be classified based on if it is full,
e.g., $q(\boldsymbol{Y|X}) = q(\text{label}|\text{feature = full}) = [0,1]$.
It can be understood that the training labels categorize the objects into two groups and DNNs can extract the hidden feature to generate $q(\boldsymbol{Y|X})$ for precisely formulating the categorization, though it is imperceptible or indecipherable for humans.
However, the categorization indicated by the training labels is obviously not consistent with the testing labels, because labels are random, thereby the testing error becomes $50\%$.
That explains why DNNs has a high testing error on random labels.
Since DNNs still can learn an accurate prior distribution $q(\boldsymbol{X})$, we conclude that this experiment does not contradict the generalization property of deep learning.

\pagebreak
\section{Conclusion}

In this work, we present a novel probabilistic representation for explaining DNNs and investigate two fundamental properties of deep learning: {hierarchy} and {generalization}. 
First, we explicitly formulate the {hierarchy} property from the Bayesian perspective.
Second, we demonstrate that DNNs have an explicit {regularization} by learning a prior distribution and clarify some empirical phenomena of DNNs that cannot be explained by traditional theories of generalization. 
Simulation results validate the proposed probabilistic representation and the insights based on a synthetic dataset.

%\pagebreak

\bibliography{neurips_2019}

\begin{thebibliography}{10}

\bibitem{Soatto1}
Alessandro Achille and Stefano Soatto.
\newblock Emergence of invariance and disentanglement in deep representations.
\newblock {\em arXiv preprint arXiv:1706.01350}, 2017.

\bibitem{VC-dimension}
Peter~L. Bartlett, Nick Harvey, Christopher Liaw, and Abbas Mehrabian.
\newblock Nearly-tight vc-dimension and pseudodimension bounds for piecewise
  linear neural networks.
\newblock {\em arXiv preprint arXiv:1703.02930}, 2017.

\bibitem{representation-dl}
Yoshua Bengio, Aaron Courville, and Pascal Vincent.
\newblock Representation learning: A review and new perspectives.
\newblock {\em IEEE Transactions on Pattern Analysis and Machine Intelligence},
  35(8):1798--1828, 2013.

\bibitem{VI}
David Blei, Alp Kucukelbir, and Jon MaAuliffe.
\newblock Variational inference: A review for statisticians.
\newblock {\em Journal of Machine Learning Research}, 112:859--877, 2017.

\bibitem{uniform_stability}
Olivier Bousquet and Andre Elisseeff.
\newblock Stability and generalization.
\newblock {\em Journal of Machine Learning Research}, pages 499--526, 2002.

\bibitem{Gibbs_softmax}
John Bridle.
\newblock Training stochastic model recognition algorithms as networks can lead
  to maximum mutual information estimation of parameters.
\newblock In {\em NeurIPS}, 1990.

\bibitem{gd_vr}
Pratik Chaudhari and Stefano Soatto.
\newblock Stochastic gradient descent performs variational inference, converges
  to limit cycles for deep networks.
\newblock In {\em ITA}, 2018.

\bibitem{it_book}
Thomas Cover and Joy Thomas.
\newblock {\em Elements of Information Theory}.
\newblock Wiley-Interscience, Hoboken, New Jersy, 2006.

\bibitem{Geman}
S.~Geman and D.~Geman.
\newblock Stochastic relaxation, gibbs distributions, and the bayesian
  restoration of images.
\newblock {\em IEEE Transactions. on Pattern Analysis and Machine
  Intelligence}, pages 721--741, June 1984.

\bibitem{posterior1}
Herbert Gish.
\newblock A probabilistic approach to the understanding and training of neural
  network classifiers.
\newblock In {\em IEEE ICASSP}, pages 1361--1364, 1990.

\bibitem{Goodfellow-et-al-2016}
Ian Goodfellow, Yoshua Bengio, and Aaron Courville.
\newblock {\em Deep Learning}.
\newblock MIT Press, 2016.

\bibitem{conjugate_gibbs}
James Hensman, Magnus Rattray, and Neil~D. Lawrence.
\newblock Fast variational inference in the conjugate exponential family.
\newblock In {\em NeurIPS}, 2012.

\bibitem{CD}
Geoffrey~E. Hinton.
\newblock Training products of experts by minimizing contrastive divergence.
\newblock {\em Neural Computation}, 14:1771--1800, 2002.

\bibitem{SVI}
Matthew~D. Hoffman, David~M. Blei, Chong Wang, and John Paisley.
\newblock Stochastic variational inference.
\newblock {\em Journal of Machine Learning Research}, 14:1303--1347, 2013.

\bibitem{ml_vi}
Michael~I. Jordan, Zoubin Ghahramani, Tommi~S. Jaakkola, and Lawrence~K. Saul.
\newblock An introduction to variational methods for graphical models.
\newblock {\em Machine Learning}, 37:183--233, 1999.

\bibitem{xinjie2}
Xinjie Lan and Kenneth~E. Barner.
\newblock From mrfs to cnns: A novel image restoration method.
\newblock In {\em 52nd Annual Conference on Information Sciences and Systems
  (CISS)}, pages 1--5, 2018.

\bibitem{energy_learning}
Yann LeCun, Sumit Chopra, Raia Hadsell, Marc'Aurelio Ranzato, and Fu~Jie Huang.
\newblock {\em A tutorial on energy-based learning}.
\newblock MIT Press, 2006.

\bibitem{BHM}
Feifei Li and Pietro Perona.
\newblock A bayesian hierarchical model for learning natural scene categories.
\newblock In {\em CVPR}, 2005.

\bibitem{MRF_IA}
Stan~Z. Li.
\newblock {\em Markov Random Field Modeling in Image Analysis 2nd ed.}
\newblock Springer, New York, 2001.

\bibitem{Boltzmann_machine}
Pankaj Mehta and David~J. Schwab.
\newblock An exact mapping between the variational renormalization group and
  deep learning.
\newblock {\em arXiv preprint arXiv:1410.3831}, 2014.

\bibitem{Conjugate}
Kevin~P. Murphy.
\newblock Conjugate bayesian analysis of the gaussian distribution.
\newblock {\em Technical report, University of British Columbia}, 2007.

\bibitem{DNN-Bottleneck}
Noga~Zaslavsky Naftali~Tishby.
\newblock Deep learning and the information bottleneck principle.
\newblock {\em arXiv preprint arXiv:1503.02406}, 2015.

\bibitem{generalization_regularization1}
Behnam Neyshabur, Srinadh Bhojanapalli, David McAllester, and Nathan Srebro.
\newblock Exploring generalization in deep learning.
\newblock In {\em NeurIPS}, 2017.

\bibitem{regularization5}
Behnam Neyshabur, Ryota Tomioka, and Nathan Srebro.
\newblock In search of the real inductive bias: On the role of implicit
  regularization in deep learning.
\newblock In {\em ICLR}, 2015.

\bibitem{dis_vs_gen}
Andrew~Y. Ng and Michael~I. Jordan.
\newblock On discriminative vs. generative classifiers: A comparison of
  logistic regression and naive bayes.
\newblock In {\em NeurIPS}, pages 841--848, 2002.

\bibitem{exponential}
Frank Nielsen and Vincent Garcia.
\newblock Statistical exponential families: A digest with flash cards.
\newblock {\em arxiv preprint arXiv:0911.4863}, 2011.

\bibitem{DRMM}
Ankit Patel, Minh Nguyen, and Richard Baraniuk.
\newblock A probabilistic framework for deep learning.
\newblock In {\em NeurIPS}, 2016.

\bibitem{pearl}
Judea Pearl.
\newblock Theoretical impediments to machine learning with seven sparks from
  the causal revolution.
\newblock {\em arXiv preprint arXiv:1801.04016}, 2018.

\bibitem{cnn-posterior}
M.D. Richard and R.P. Lippmann.
\newblock Neural network classifiers estimate bayesian a posteriori
  probabilities.
\newblock {\em Neural Computation}, pages 461--483, 1991.

\bibitem{backpropagation}
David~E. Rumelhart, Geoffrey~E. Hinton, and Ronald~J. Williams.
\newblock Learning representations by back-propagating errors.
\newblock {\em Nature}, 323:533--536, October 1986.

\bibitem{deep-BM}
Ruslan Salakhutdinov and Geoffrey Hinton.
\newblock Deep boltzmann machines.
\newblock In {\em AISTATS 2009}, pages 448--455, 2009.

\bibitem{IP-argue}
Andrew Saxe, Yamini Bansal, Joel Dapello, Madhu Advani, Artemy Kolchinsky,
  Brendan Tracey, and David Cox.
\newblock On the information bottleneck theory of deep learning.
\newblock In {\em ICLR}, 2018.

\bibitem{DNN-information}
Ravid Shwartz-Ziv and Naftali Tishby.
\newblock Opening the black box of deep neural networks via information.
\newblock {\em arXiv preprint arXiv:1703.00810}, 2017.

\bibitem{statistical_img}
E.~P. Simoncelli.
\newblock Statistical models for images: Compression, restoration and
  synthesis.
\newblock In {\em Proc 31st Asilomar Conf on Signals, Systems and Computers},
  pages 673--678, November 1997.

\bibitem{bayesian_regularization}
Harald Steck and Tommi~S. Jaakkola.
\newblock On the dirichlet prior and bayesian regularization.
\newblock In {\em NeurIPS}, 2003.

\bibitem{DMFA}
Yichuan Tang, Ruslan Salakhutdinov, and Geoffrey Hinton.
\newblock Deep mixtures of factor analysers.
\newblock {\em arXiv preprint arXiv:1206.4635}, 2015.

\bibitem{GSM-stat}
Martin.~J. Wainwright and Eero.~P. Simoncelli.
\newblock Scale mixtures of gaussians and the statistics of natural images.
\newblock In {\em NeurIPS}, pages 855--861, 2000.

\bibitem{generalization_regularization}
Chiyuan Zhang, Samy Bengio, Moritz Hardt, Benjamin Recht, and Oriol Vinyals.
\newblock Understanding deep learning requires rethinking generalization.
\newblock In {\em ICLR}, 2016.

\bibitem{posterior4}
G.~Zhang.
\newblock Neural networks for classification: a survey.
\newblock {\em IEEE Transactions on Systems, Man, and Cybernetics},
  30:451--462, 2000.

\bibitem{CRF_RNN}
Shuai Zheng, Sadeep Jayasumana, Bernardino Romera-Paredes, Vibhav Vineet,
  Zhizhong Su, Dalong Du, Chang Huang, and Philip Torr.
\newblock Conditional random fields as recurrent neural networks.
\newblock In {\em International Conference on Computer Vision (ICCV)}, pages
  1529--1537, 2015.

\end{thebibliography}

%[1] Alexander, J.A.\ \& Mozer, M.C.\ (1995) Template-based algorithms for
%connectionist rule extraction. In G.\ Tesauro, D.S.\ Touretzky and T.K.\ Leen
%(eds.), {\it Advances in Neural Information Processing Systems 7},
%pp.\ 609--616. Cambridge, MA: MIT Press.

%[2] Bower, J.M.\ \& Beeman, D.\ (1995) {\it The Book of GENESIS: Exploring
%  Realistic Neural Models with the GEneral NEural SImulation System.}  New York:
%TELOS/Springer--Verlag.

%[3] Hasselmo, M.E., Schnell, E.\ \& Barkai, E.\ (1995) Dynamics of learning and
%recall at excitatory recurrent synapses and cholinergic modulation in rat
%hippocampal region CA3. {\it Journal of Neuroscience} {\bf 15}(7):5249-5262.

\end{document}